\documentclass[letterpaper]{article} 
\usepackage{aaai2026}  
\usepackage{times}  
\usepackage{helvet}  
\usepackage{courier}  
\usepackage[hyphens]{url}  
\usepackage{graphicx} 
\usepackage{natbib}  
\usepackage{caption} 
\usepackage{algorithm}
\usepackage{algorithmic}
\usepackage{amsmath}
\usepackage{amssymb}
\usepackage{multirow}
\usepackage{bm}
\usepackage{booktabs}
\usepackage{microtype}
\usepackage{pifont} 
\usepackage{float}
\usepackage{caption} 
\usepackage{subcaption} 

\usepackage{newfloat}
\usepackage{listings}
\DeclareCaptionStyle{ruled}{labelfont=normalfont,labelsep=colon,strut=off} 
\floatstyle{ruled}
\newfloat{listing}{tb}{lst}{}
\floatname{listing}{Listing}
\nocopyright 

\newcommand{\name}{StreamKV }

\title{\name: Streaming Video Question-Answering with Segment-based KV Cache Retrieval and Compression
}

\author{
    Yilong Chen\textsuperscript{\rm 1,\rm 2},
    Xiang Bai\textsuperscript{\rm 1,\rm 2},
    Zhibin Wang\footnote{Project Leader}\textsuperscript{\rm 2},
    Chengyu Bai\textsuperscript{\rm 1},
    Yuhan Dai\textsuperscript{\rm 2},
    Ming Lu\textsuperscript{\rm 1}\\
    Shanghang Zhang\footnote{Corresponding Author}\textsuperscript{\rm 1}
}
\affiliations{
    \textsuperscript{\rm 1}State Key Laboratory of Multimedia Information Processing, School of Computer Science, Peking University\\
    \textsuperscript{\rm 2}TaoBao \& Tmall Group of Alibaba\\
    chenyl@stu.pku.edu.cn
}

\usepackage{bibentry}
\usepackage{bbding}

\begin{document}

\maketitle

\begin{abstract}
Video Large Language Models (Video-LLMs) have demonstrated significant potential in the areas of video captioning, search, and summarization. 
However, current Video-LLMs still face challenges with long real-world videos. 
Recent methods have introduced a retrieval mechanism that retrieves query-relevant KV caches for question answering, enhancing the efficiency and accuracy of long real-world videos. 
However, the compression and retrieval of KV caches are still not fully explored. 
In this paper, we propose \textbf{StreamKV}, a training-free framework that seamlessly equips Video-LLMs with advanced KV cache retrieval and compression.
Compared to previous methods that used uniform partitioning, StreamKV dynamically partitions video streams into semantic segments, which better preserves semantic information. 
For KV cache retrieval, StreamKV calculates a summary vector for each segment to retain segment-level information essential for retrieval. 
For KV cache compression, StreamKV introduces a guidance prompt designed to capture the key semantic elements within each segment, ensuring only the most informative KV caches are retained for answering questions.
Moreover, StreamKV unifies KV cache retrieval and compression within a single module, performing both in a layer-adaptive manner, thereby further improving the effectiveness of streaming video question answering. 
Extensive experiments on public StreamingVQA benchmarks demonstrate that StreamKV significantly outperforms existing Online Video-LLMs, achieving superior accuracy while substantially improving both memory efficiency and computational latency. 
The code has been released at https://github.com/sou1p0wer/StreamKV.
\end{abstract}

\section{Introduction}
\label{sec:paper-introduction}

\begin{figure}[t]
    \centering
    \includegraphics[width=0.95\columnwidth]{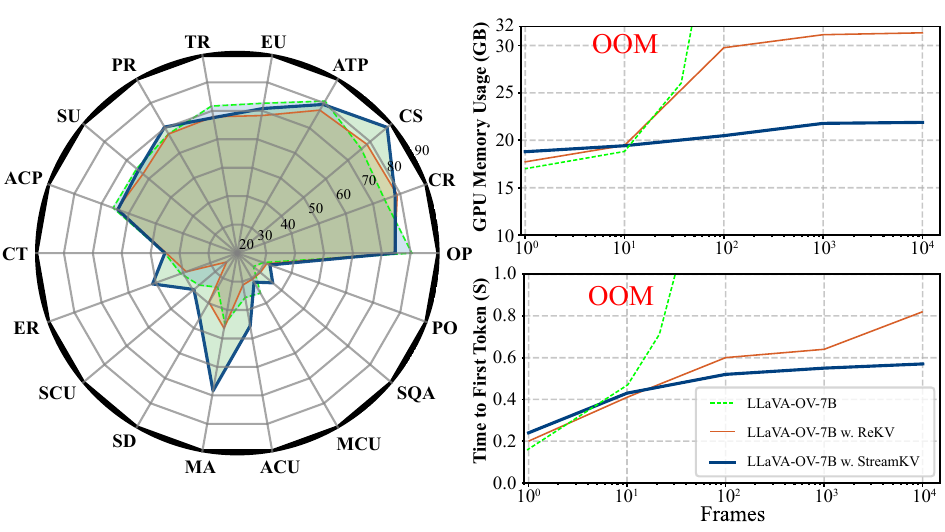}
    \caption{Comparison with ReKV on StreamingBench in terms of VideoQA accuracy, memory usage, and latency}
    \label{fig:performance}
\end{figure}

Recent advancements in large language models (LLMs)\cite{touvron2023llama,touvron2023llama2,ouyang2022instructgpt,openai2023gpt4,bai2023qwen} have significantly enhanced the capabilities of Video-LLMs across various video understanding tasks, such as video captioning\cite{qasim2023densevideocaptioningsurvey}, search\cite{luo2021clip4clipempiricalstudyclip}, and summarization\cite{alaa2024videosummarizationtechniquescomprehensive}. 
Although significant progress has been made, most current Video-LLMs are primarily designed for offline usage. The model processes the entire video along with all questions at once, which leads to substantial GPU memory usage and increased response latency as video lengths grow.
This paradigm inherently restricts their applicability to relatively short video clips and fails to meet the requirements of real-world, real-time interactive scenarios such as autonomous driving\cite{cao2025fastdrivevlaefficientendtoenddriving}, embodied AI\cite{li2025manipdreamer3dsynthesizingplausible}, and augmented reality (AR) devices, where video streams continuously and user queries arrive on-the-fly.

Several pioneering studies have explored the paradigm of online Video-LLMs, where models continuously process video streams and answer user queries based on previous observations. Online Video-LLMs involve three key challenges: (1) How to efficiently and effectively process streaming video; (2) How to balance the retention of visual context with memory consumption; and (3) How to quickly and accurately retrieve relevant historical information when responding to user queries. 

Recently, ReKV~\cite{di2025rekv} has introduced a retrieval mechanism that retrieves only query-relevant KV-caches, enhancing the efficiency and accuracy of question answering. However, this approach divides the video stream into uniform segments, disrupting the continuous semantic structure of the video and storing the entire historical visual context, 
leading to significant memory consumption. Moreover, its retrieval mechanism remains inflexible.

To this end, we propose StreamKV, a training-free framework that seamlessly equips Video-LLMs with advanced KV cache retrieval and compression.
Unlike previous methods that uniformly segment video streams and risk arbitrarily breaking semantic boundaries, StreamKV dynamically partitions the video stream into semantic segments, which better preserves semantic information. 
For KV cache retrieval, StreamKV calculates a summary vector for each segment to retain segment-level information. The KV pairs generated from this summary vector are excluded from further compression, which is crucial for accurately answering related questions.
For KV cache compression, existing methods mainly target the decoding stage to reduce memory and computational overhead during inference. Although effective in offline settings, they are not suitable for StreamingVQA. In contrast, our approach treats semantic segments as the basic unit of compression and applies compression immediately after each segment is encoded. In the StreamingVQA scenario, since user questions are typically unknown when performing KV compression and multi-turn dialogues are expected, the compression process should focus on video semantics rather than specific user questions. To achieve this, StreamKV introduces a guidance prompt to capture the key semantic elements within each segment, ensuring only the most informative KV caches are retained for answering questions, thereby effectively removing redundancy.
To further enhance both KV compression and retrieval, StreamKV introduces a unified layer-adaptive KV selection module, which dynamically allocates selection budgets across all transformer layers under an overall budget. This approach leverages the distinct information distributions in transformer layers instead of uniform allocation. This allows for optimal budget allocation, increasing the overall informative content retained under a fixed total budget.

We compare StreamKV with the latest method, ReKV~\cite{di2025rekv}, on StreamingBench~\cite{lin2024streamingbenchassessinggapmllms} using the same foundation model. As shown in Figure~\ref{fig:performance}, StreamKV outperforms ReKV in terms of VideoQA accuracy, memory usage, and latency.

In summary, our main contributions are as follows:
\begin{itemize}
\item We propose StreamKV,  a training-free framework that seamlessly equips Video-LLMs with advanced KV cache retrieval and compression.
\item To better preserve the semantic continuity of video content, StreamKV adopts a semantic partitioning and summary vector mechanism. This approach facilitates both subsequent compression and retrieval.
\item To enable KV cache compression in streaming scenarios, we introduce a guidance prompt to capture key semantic elements within each segment, ensuring essential information is retained even under aggressive compression.
\item To further improve KV cache retrieval and compression, we propose a Unified Layer-Adaptive KV Selection Module that allocates the selection budget optimally across transformer layers, maximizing informative content under a fixed total budget.
\item Comprehensive experiments demonstrate that StreamKV significantly outperforms existing Online Video-LLMs, achieving superior accuracy while substantially improving both memory efficiency and latency. 
\end{itemize}

\section{Related Works}
\label{sec:paper-related-works}
\noindent \textbf{Video Large Language Models}
With the rapid progress of Multimodal Large Language Models (MLLMs)~\cite{li2023blip2bl,Qwen2VL,li2024llavaonevision,zhang2024mm1}, Video LLMs~\cite{video-llama,video-llava,li2024videochat-flash,zhang2025videollama} have drawn great attention in recent years.
Typically, these models use a visual encoder~\cite{clip,siglip,tschannen2025siglip2multilingualvisionlanguage} to extract video features, followed by a modality projector (e.g., MLP~\cite{liu2023llava} and Q-former~\cite{li2023blip2bl}) to map visual features into language space. Then, the mapped features are combined with text tokens as input to the LLM to generate a contextual response.
While exhibiting strong performance on offline video understanding benchmarks~\cite{zhou2024mlvu,videomme}, these models are intrinsically not well-suited for streaming video understanding owing to the memory bottleneck and complexity of information in long videos\cite{cao2025movekdknowledgedistillationvlms}.
To bridge this gap, our work seamlessly equips Video-LLMs with streaming capabilities.

\noindent \textbf{Streaming Video Understanding}
Streaming video understanding~\cite{wu2024videollmmodefficientvideolanguagestreaming,li2025lionfsfastslow,ding2025streammind,liu2025streamchatchattingstreamingvideo} requires Video LLMs to process real-time video frames and answer user questions based on all content up to a specified timestamp.
VideoLLM-Online~\cite{videollm_online} proposes the LIVE framework for streaming dialogue. However, it does not provide an effective solution for long-term video input processing. Subsequent works such as Flash-Vstream~\cite{vstream} and Dispider~\cite{qian2025dispider} focus on managing complex video content and improving model performance through specially designed memory-augmented architectures. Recently, ReKV~\cite{di2025rekv} introduced a retrieval mechanism that retrieves query-relevant KV caches for question answering. However, it retains all generated KV caches, resulting in substantial memory consumption, and its retrieval strategy requires further optimization. To overcome these limitations, we propose StreamKV, which substantially improves memory efficiency and retrieval effectiveness, achieving strong performance in streaming video understanding.

\noindent \textbf{KV Cache Compression for Video LLMs}
Efficient KV cache compression~\cite{cai2025pyramidkvdynamickvcache,yang2024pyramidinferpyramidkvcache,wang2025sparsemmheadsparsityemerges} is essential in MLLMs to manage memory and latency overhead. FastV~\cite{chen2024image} accelerates the prefill phase by pruning visual tokens in specific layers, utilizing attention scores from the final query token to guide the selection. Similarly, SparseVLM~\cite{zhang2024sparsevlm} employs cross-attention to identify visual tokens relevant to user queries. However, most existing methods~\cite{li2024snapkv,zhang2023ho,fu2024headkv} are heavily based on a given user question, limiting their robustness and applicability in StreamingVQA scenarios. To address this challenge, StreamKV introduces a guidance prompt to capture the key semantic elements within each segment rather than relying on specific user questions.

\section{Method}
\label{sec:paper-method}

\begin{figure*}[t]
\centering
\includegraphics[width=\textwidth]{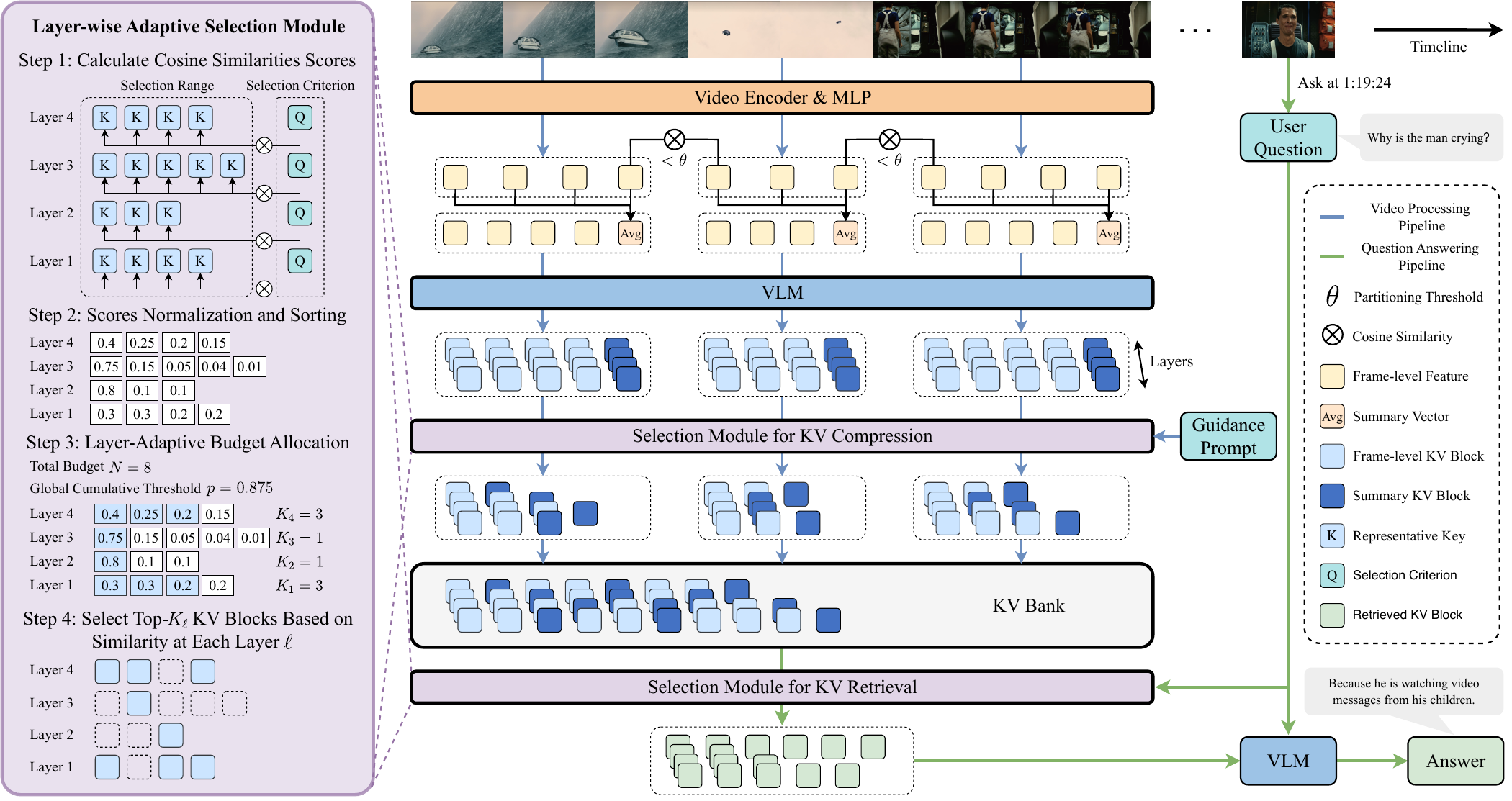}
\caption{StreamKV workflow. StreamKV dynamically partitions video streams into semantic segments, and calculates a summary vector for each segment. These segments are sequentially encoded to generate frame-level KV blocks. KV compression is applied to all blocks except for those generated by the summary vector. We store the compressed KV blocks and summary KV block in the KV Bank. Upon receiving a question, StreamKV retrieves query-relevant KV blocks from the KV Bank to generate responses. Both the compression and retrieval are performed using our proposed unified layer-adaptive selection module.}
\label{overview}
\end{figure*}

We first introduce the workflow of StreamKV, as illustrated in Figure~\ref{overview}. StreamKV partitions the video stream into semantic segments and calculates a summary vector for each segment. These segments are sequentially encoded to generate frame-level KV blocks, which are then compressed - except for those derived from the summary vector - and stored in the KV Bank. Upon receiving a question, StreamKV retrieves query-relevant KV blocks from the KV Bank to generate responses. Both the compression and retrieval of KV blocks are performed using our proposed unified layer-adaptive selection module.

\subsection{Semantic Segment Partitioning and Encoding} 
\begin{figure}[t]
    \centering
    \begin{subfigure}[b]{0.48\linewidth}
        \includegraphics[width=\linewidth]{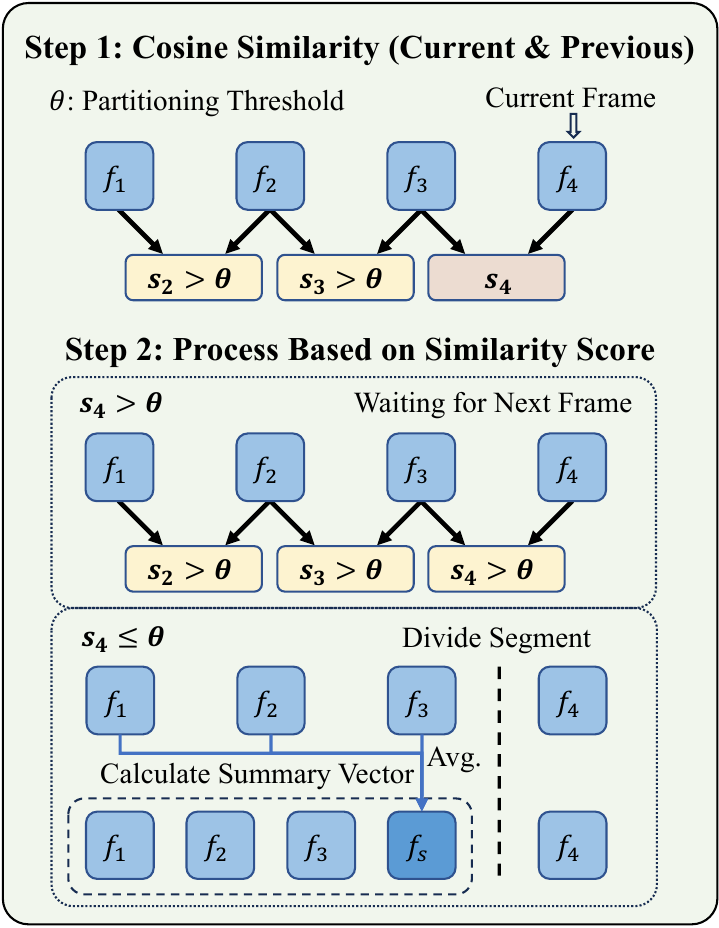} 
        \caption{Semantic Partitioning}
        \label{fig:seg}
    \end{subfigure}
    \hfill
    \begin{subfigure}[b]{0.48\linewidth}
        \includegraphics[width=\linewidth]{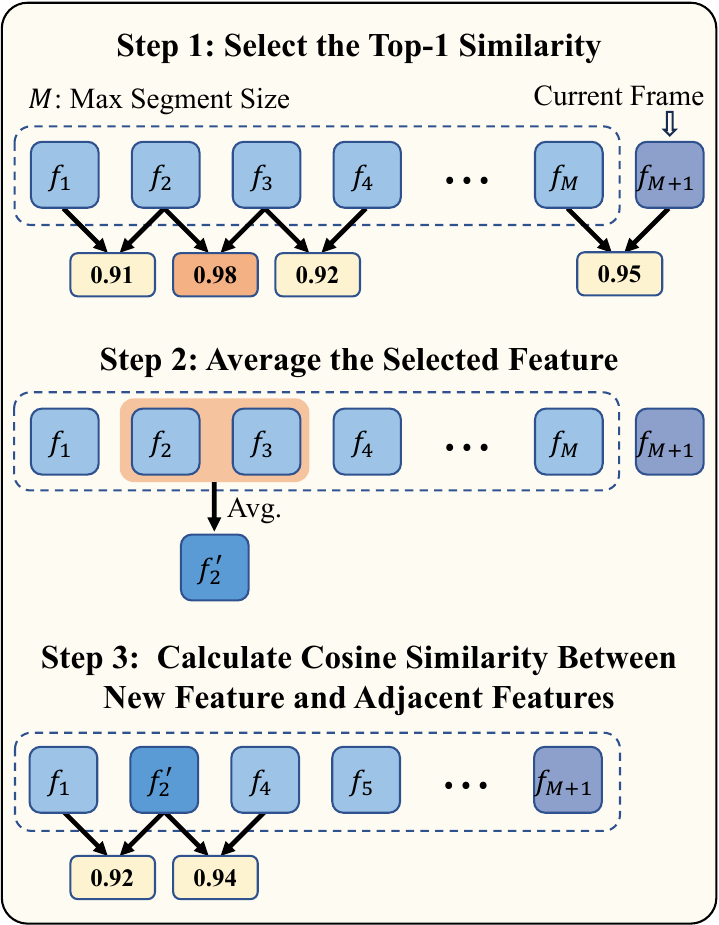}
        \caption{Segment Merging}
        \label{fig:merge}
    \end{subfigure}
    \caption{Video Segment Processing}
    \label{Video Segment Processing}
\end{figure}

\noindent \textbf{Semantic Segment Partitioning}
We sample video frames at a regular interval and extract embedding $f_t\in \mathbb{R}^{ P^{2}  \times D}$ for each frame using a vision encoder, where $P^{2}$ is the number of ViT patch tokens and $D$ is the hidden dimension of ViT~\cite{dosovitskiy2021imageworth16x16words}. 
To detect significant visual changes and potential semantic boundaries, we compute the cosine similarity between adjacent embeddings:
\begin{equation}
s_t = {\textstyle \frac{f_{t-1} \cdot f_t}{\|f_{t-1}\|\, \|f_t\|}}.
\end{equation}
As illustrated in Figure~\ref{fig:seg}, frames with low similarity scores are identified as semantic boundaries, each indicating the start of a new segment. 

To avoid excessively short segments, we apply an exclusion window of size $m$ around each boundary, ensuring that the resulting segments are of sufficient length to contain relevant information. To limit segment length, we introduce a segment merging technique to exploit the temporal redundancies inherent in videos, as illustrated in Figure~\ref{fig:merge} . If the current segment exceeds a threshold $M$, we merge the most similar adjacent frame pair, based on previously computed cosine similarities~\cite{xu2024slowfastllavastrongtrainingfreebaseline,chatunivi}.

Through semantic partitioning, the video stream is continually divided into a sequence of semantic segments $ \left [ \mathbf{S}^i \right ]  $. Each segment is defined as $\mathbf{S}^i := [f_t^i]_{t=1}^{T_i}$, where the segment length $T_{i} $ satisfies $T_i \in [m, M]$.
To preserve segment-level information, we calculate a summary vector  $f_{s}^{i} = \frac{1}{T_i} {\textstyle\sum_{t=1}^{T_i}} f_{t}^{i}$  for each segment by averaging frame-level features at each spatial location.
\\\\
\noindent \textbf{Segment-based Sliding-window Encoding}
We encode the video stream $\mathcal{V}^T$ incrementally, processing it segment by segment.
At each step, given the current segment $\mathbf{S}^i$, its summary vector $f_{s}^{i} $ , and a local window of past KV pairs $\mathbf{L}$, we compute attention~\cite{vaswani2017attention}
\begin{equation}
\mathbf{O} = \mathrm{Attn}\big(\mathbf{W_Q}\mathbf{X}^i,\, [\mathbf{L}_k,\, \mathbf{W_K}\mathbf{X}^i],\, [\mathbf{L}_v,\, \mathbf{W_V}\mathbf{X}^i]\big),
\label{equ:video_encoding}
\end{equation}
where $\mathbf{W_Q}$, $\mathbf{W_K}$, and $\mathbf{W_V}$ are the attention parameters,  $\mathbf{L}_k$ and $\mathbf{L}_v$ denote the key and value vectors of $\mathbf{L}$, and $\mathbf{X}^i=[\mathbf{S}^i \Vert f_{s}^{i}]$ is the concatenation of $\mathbf{S}^i$ and $f_{s}^{i}$.

For each segment $\mathbf{S}^i$, we derive a collection of frame-level KV blocks. Specifically, the $m$-th KV block is defined as $\mathbf{b}_m^i = [ (\mathbf{k}_{m,p}^i,\, \mathbf{v}_{m,p}^i ) ]_{p=1}^{P^2}$, where $\mathbf{k}_{m,p}^i,\, \mathbf{v}_{m,p}^i $ denote the patch-level key and value vectors of the $m$-th frame. The representative key vector for  $\mathbf{b}_m^i$  is computed by averaging its patch-wise key vectors: 
\begin{equation}
\mathbf{r}_m^i = \frac{1}{P^2} {\textstyle\sum_{p=1}^{P^2}} \mathbf{k}_{m,p}^i \in \mathbb{R}^{D'}.
\end{equation}
For computational efficiency, we do not distinguish between attention heads and instead concatenate them into a single vector of dimension $D'$.
The summary vector $f_{s}^{i}$ similarly yields a summary KV block $ \mathbf{b}_s^i$ and representative key $\mathbf{r}_s^i $.

We extend this notation to all \textit{L} network layers: for each segment \textit{i} and each layer $l$, we denote the frame-level KV blocks and representative key vectors as $\mathbf{b}_m^{i,l}  $ and $\mathbf{r}_m^{i,l}$ respectively. Likewise, the summary KV block and its representative key vector are denoted as $\mathbf{b}_s^{i,l}$ and $\mathbf{r}_s^{i,l}$.
At each layer $l$, the collection of KV blocks and their representative keys for segment $i$ are given by:
\begin{equation}
    \bm{B}_l^{i} := [ \mathbf{b}_m^{i,l} ]_{m=1}^{T_{i}}, \quad
    \bm{R}_l^{i} := [ \mathbf{r}_m^{i,l} ]_{m=1}^{T_{i}}
    \label{eq:R}.
\end{equation}
Here,  $\mathbf{b}_s^{i,l}$ is specifically preserved to retain segment-level information, while $ \mathcal{B}_l^{i}$  is subject to further compression.

\subsection{Unified Layer-Adaptive KV Selection Module}

We formulate both KV compression and retrieval as the problem of selecting the most relevant KV entries from a selection range according to a specific selection criterion~\cite{lewis2021retrievalaugmentedgenerationknowledgeintensivenlp}. Specifically, for compression, we use a guidance prompt to select the most informative KV blocks within a segment; for retrieval, we use user questions to select query-relevant KV blocks from the KV Bank.
To efficiently address this, we introduce a Unified Layer-Adaptive KV Selection Module. 
Unlike uniform allocation strategies, this approach adaptively allocates selection budgets across transformer layers according to their information distributions, thereby increasing the overall retained content under a fixed budget.

\noindent \textbf{Step-1: Calculate Cosine Similarity Scores}
Given $L$ transformer layers, for each layer $l$, let $\bm{R}_l$ be the selection range, i.e., a sequence of candidate representative key vectors $[ \mathbf{r}_j^l ]_{j \in \operatorname{idx}(\bm{R}_l)}
$, where $\operatorname{idx}(\bm{R}_l)$ denotes the indices of $\bm{R}_l$, and let $\mathbf{c}^l$ represent the selection criterion vector for layer $l$. We define $\mathrm{Sim}_l(j)$ as the cosine similarity between each candidate $\mathbf{r}_j^l$ and the criterion $\mathbf{c}^l$.

The selection problem is thus formulated as selecting an index subset $\mathcal{I}_l$ for each layer, corresponding to the top $K_l$ candidates with the highest similarity,
\begin{equation}
    \mathcal{I}_l = \operatorname{Top}K_{\text{idx}} \bigl( [ \mathrm{Sim}_l(j) ]_{j \in \operatorname{idx}(\bm{R}_l)},\; K_l \bigr),
\end{equation}
where $K_l\le  \left | \bm{R}_l \right |  $ and ${\textstyle \sum_{l=1}^{L}}K_l =N$. Here, $ \left |  \bm{R}_l \right |  $ is the size of $\bm{R}_l$ and $N$ denotes the total selection budget.

In contrast to uniform allocation, we adaptively assign $K_l$ based on the similarity distribution in each layer, increasing the total informative content retained under the budget $N$.



\noindent \textbf{Step-2: Scores Normalization and Sorting}
For each layer $l$, we first calculate its normalized similarity score distribution~\cite{rumelhart1986learning}:
\begin{equation}
\widetilde{\mathrm{Sim}}_l(j) = 
\frac{\exp(\mathrm{Sim}_l(j))}{\sum_{k=1}^{| \bm{R}_l |} \exp(\mathrm{Sim}_l(k))},
\quad j \in \operatorname{idx}(\bm{R}_l).
\end{equation}
We then sort the normalized scores in descending order to derive the priority sequence $[\widetilde{\mathrm{Sim}}_l(s_l(j))]_{j \in \operatorname{idx}(\bm{R}_l)}$, where $s_l(j)$ represents the index of the $j$-th largest score in $\bm{R}_l$.

\noindent \textbf{Step-3: Layer-Adaptive Budget Allocation}
Let $p$ be a global cumulative score threshold, which serves as an intermediate variable to facilitate the determination of the allocation. For each layer $l$, we define $K_{l}^{p}$ as the minimal prefix length such that the cumulative sum of the top $K_{l}^{p}$ normalized scores reaches or exceeds $p$:
\begin{equation}
K_{l}^{p} = \min\{ k \mid {\textstyle \sum_{j=1}^{k}} \widetilde{\mathrm{Sim}}_l(s_l(j)) \geq p \}.
\end{equation}
The global threshold \textit{p} is determined such that the sum of selected candidates across all layers satisfies the total selection budget constraint, i.e.,${\textstyle \sum_{l=1}^{L}}K_{l}^{p} =N$. This can be efficiently solved via a binary search over possible $p$ values. Once the appropriate $p$ is identified, we obtain the corresponding allocation $\{ K_l^p \}_{l=1}^{L}$, which serves as the final budget allocation $\{ K_l \}_{l=1}^{L} $. Finally, based on this budget allocation, we derive the selected index subsets $\{ \mathcal{I}_l \}_{l=1}^{L}
$ for each layer. 
\begin{algorithm}[t]
\caption{Binary Search for Global Threshold $p$}
\label{algo:binary-search}
\begin{algorithmic}[1]
\STATE \textbf{Input:} Total budget $N$,
\STATE \hspace{2.8em} Priority sequence $[\widetilde{\mathrm{Sim}}_l(s_l(j))]_{j \in \operatorname{idx}(\mathcal{R}_l)},\ \forall l$ 
\STATE Initialize $p_1 \gets 0$, $p_2 \gets 1$
\WHILE{$p_2 - p_1 > \epsilon$}
    \STATE $p \gets \frac{p_1 + p_2}{2}$
    \STATE $K_{l}^{p} = \min\{\, k \mid \sum_{j=1}^{k} \widetilde{\mathrm{Sim}}_l(s_l(j)) \geq p \},
\quad \forall l$
    \STATE $\delta \gets \sum_l K_{l}^{p} - N$
    \STATE \textbf{if} $\delta = 0$ \textbf{then} \textbf{return} $p$
    \STATE \textbf{else if} $\delta < 0$ \textbf{then} $p_1 \gets p$
    \STATE \textbf{else} $p_2 \gets p$
\ENDWHILE
\STATE \textbf{return} $p$
\end{algorithmic}
\end{algorithm}

In summary, we encapsulate the above procedure into a unified selection function:
\begin{equation}
\{ \mathcal{I}_l \}_{l=1}^L = \mathrm{SelectKV} ( \{ \mathcal{R}_l, \mathbf{c}^l \}_{l=1}^L, N ),
\end{equation}
where SelectKV denotes the layer-adaptive KV selection module, with inputs $\{ \mathcal{R}_l, \mathbf{c}^l \}_{l=1}^L$ (i.e., the candidate representative KV vectors and the corresponding selection criteria for each layer) and the total selection budget $N$. The output is the set of selected representative indices $\left\{  \mathcal{I}_l \right\}_{l=1}^L$.

\subsection{\textbf{KV Compression via Guidance Prompt}}

In StreamingVQA, video segments are encoded sequentially, and the resulting KV caches are stored for subsequent tasks. As input length increases, storing all KV caches becomes infeasible. Existing KV compression methods mainly target the decoding stage to reduce memory and computational overhead during inference. Although effective in offline settings, they are not suitable for StreamingVQA. In contrast, our compression method is applied immediately after each segment is encoded, retaining only the most informative KV caches from each segment in our KV bank, significantly reducing memory consumption. Moreover, since user questions are typically unknown when performing KV compression and multi-turn dialogues~\cite{zhu2023survey} are expected, the compression process should focus on video semantics rather than specific user questions.

To address these requirements, StreamKV introduces a guidance prompt designed to capture the key semantic elements within each segment, such as \textbf{salient entities} (e.g., people, objects, locations, key visual concepts), \textbf{key events and actions} (what happened, when, and where), \textbf{temporal and causal relationships} (how events unfold and cause-effect chains), \textbf{contextual cues} (scene changes, dialogue, narrative shifts) and \textbf{important numerical or factual details} (for tasks like counting, summarization, or fact-based QA). This approach enables segments to preserve comprehensive visual semantics and contextual coherence, thereby ensuring that essential information is retained even under aggressive compression. For reference, we present examples of our guidance prompts in Appendix A.


For each segment, we use the unified layer-adaptive KV selection module to compress redundant KV blocks. 
Specifically, for each layer $l$, the selection range is the set of representative key vectors $\bm{R}_l^{i} $ (as defined in Eq.~\eqref{eq:R}), and the selection criterion is the guidance prompt vector $\mathbf{g}^l = \frac{1}{N_g} \sum_{k=1}^{N_g} \mathbf{g}_k^l ,$ where $N_g$ is the number of tokens in the guidance prompt, and $\mathbf{g}_k^l$ is the $k$-th query vector for layer $l$. Here, $\mathbf{g}^l\in \mathbb{R}^{D'}$,  where ${D'}$ is the dimension of both the guidance vector and the representative key vectors in $\bm{R}_l^{i} $.
Given the compression ratio $\theta $,  the total selection budget $N= \left \lceil\left ( 1- \theta\right ){T_{i} }  \right \rceil\times L$ , where $T_{i}$ is the frame count of segment $i$ and $L$ is the number of layers.
The indices of the most informative KV blocks for each layer are selected as:
\begin{equation}
\textstyle
\{ \mathcal{I}_l^i \}_{l=1}^L = \mathrm{SelectKV} ( \{ \bm{R}_l^{i}, \mathbf{g}^l \}_{l=1}^L,\, N ).
\end{equation}
The compressed frame-level KV blocks and their representative key vectors for each segment and layer are then constructed as \(\tilde{\bm{B}}_l^{i} = [\, \mathbf{b}_m^{i,l} \mid m \in \mathcal{I}_l^i \,]\) and \(\tilde{\bm{R}}_l^{i} = [\, \mathbf{r}_m^{i,l} \mid m \in \mathcal{I}_l^i \,]\), where \([\,\cdot\,]\) denotes ordered concatenation.
To preserve segment-level information, we also explicitly retain the segment summary KV block  $\mathbf{b}_s^{i,l}$. 
For each layer $l$, the KV block bank $  \mathcal{B}_l$ and the set of its corresponding representative key vectors $\mathcal{R}_l$ are continuously updated in parallel:
\begin{equation}
    \mathcal{B}_l \gets [\, \mathcal{B}_l,\, \tilde{\bm{B}}_l^{i},\, \mathbf{b}_s^{i,l} \,], \quad
    \mathcal{R}_l \gets [\, \mathcal{R}_l,\, \tilde{\bm{R}}_l^{i},\, \mathbf{r}_s^{i,l} \,],
\end{equation}
where $\tilde{\bm{B}}_l^{i}$ and $\mathbf{b}_s^{i,l}$ are the compressed KV blocks and the segment summary block, respectively, and $\tilde{\bm{R}}_l^{i}$ and $\mathbf{r}_s^{i,l} $ are their corresponding representative key vectors.

\subsection{Streaming Video Question-Answering}

\noindent \textbf{KV Retrieval} 
Upon receiving a user question, we use the layer-adaptive KV Selection Module to retrieve query-relevant KV-caches.
For each layer $l$, the selection range is the set of representative key vectors $\mathcal{R}_l $ , and the selection criterion is the question vector $\mathbf{q}^l = {\textstyle \frac{1}{N_q} \sum_{k=1}^{N_q} \mathbf{q}_k^l} \in \mathbb{R}^{D'}
$, where $N_q$ is the number of tokens in the question and $\mathbf{q}_k^l$ is the $k$-th query vector at layer $l$. The total selection budget is set to $N = N_r\times L$, where $N_r$  is the desired average number of KV blocks to retrieve for each layer.
The indices of the retrieved KV blocks for each layer are obtained as:
\begin{equation}
\{ \mathcal{I}_l \}_{l=1}^L = \mathrm{SelectKV} ( \{ \mathcal{R}_l, \mathbf{q}^l \}_{l=1}^L,\, N ).
\end{equation}
The retrieved KV blocks are aggregated as:
\begin{equation}
    \mathcal{P}_l = \left[\, \mathcal{B}_l\left [ j \right ]  \mid j \in \mathcal{I}_l \,\right],
\end{equation}
where $\mathcal{P}_l$ denotes the set of KV blocks retrieved from the layer-$l$ KV block bank $\mathcal{B}_l$, with $\mathcal{B}_l\left [ j \right ]$ representing the $j$-th block.
The collection $\left \{ \mathcal{P}_l \right \} _{l=1}^{L} $ is used for subsequent question answering.

\noindent\textbf{Question-Answering Using Retrieved KV} 
The retrieved Video KV-caches $\left \{ \mathcal{P}_l \right \} _{l=1}^{L} $ serve as the context for video question-answering. Formally, the attention calculation is formulated as:
\begin{equation}
    \mathbf{O} = \text{Attn}\left(\mathbf{W_Q}\mathbf{X}, [\mathbf{C}_k, \mathbf{W_K}\mathbf{X}], [\mathbf{C}_v, \mathbf{W_V}\mathbf{X}]\right),
\end{equation} 
where $\mathbf{X}$ represents either the question tokens or the current token being decoded, and $\mathbf{C}_k$ and $\mathbf{C}_v$ are the key and value vectors from the context, which includes the retrieved KV caches$\left \{ \mathcal{P}_l \right \} _{l=1}^{L} $, question, and previously generated tokens. 

\noindent\textbf{Positional Encoding} 
Rotary Position Embeddings (RoPE)\cite{su2023roformerenhancedtransformerrotary} are widely adopted in Video LLMs for temporal encoding, but their effectiveness degrades in long sequences due to suppressed attention between distant tokens. To alleviate this, StreamKV employs distinct positional encoding for video encoding and question answering. For segment encoding, inspired by LM-Infinite~\cite{han2024lminfinite}, RoPE is applied only within the local window. For question answering, StreamKV treats retrieved tokens as consecutive and applies RoPE based on their relative positions rather than absolute positions. 

\section{Experiments}
\label{sec:paper-experiments}
\begin{table*}[t]
\small
\centering
\renewcommand{\arraystretch}{1.05}
\setlength{\tabcolsep}{1pt}
\begin{tabular}{
@{}p{3cm}@{}p{1.1cm}
|ccccccccccc|ccccc|ccccc|c}
\toprule
\multirow{2}{*}{\textbf{Model}} & \multirow{2}{*}{\textbf{Frames}} &
\multicolumn{11}{c|}{\textbf{Real-Time}} &
\multicolumn{5}{c|}{\textbf{Omni-Source}} &
\multicolumn{5}{c|}{\textbf{Contextual}} &
\multirow{2}{*}{\rotatebox{90}{\textbf{Overall}}} \\ 

\cmidrule(lr){3-13} \cmidrule(lr){14-18} \cmidrule(lr){19-23}
 & & \rotatebox{90}{OP}  & \rotatebox{90}{CR}  & \rotatebox{90}{CS}  & \rotatebox{90}{ATP}
   & \rotatebox{90}{EU}  & \rotatebox{90}{TR}  & \rotatebox{90}{PR}  & \rotatebox{90}{SU}
   & \rotatebox{90}{ACP} & \rotatebox{90}{CT}  & \rotatebox{90}{All}
   & \rotatebox{90}{ER}  & \rotatebox{90}{SCU} & \rotatebox{90}{SD}
   & \rotatebox{90}{MA}  & \rotatebox{90}{All}
   & \rotatebox{90}{ACU} & \rotatebox{90}{MCU} & \rotatebox{90}{SQA}
   & \rotatebox{90}{PO}  & \rotatebox{90}{All}
   & 
\\
\midrule
\multicolumn{24}{c}{Proprietary MLLMs} \\
\midrule
Gemini1.5 pro     & 1 fps & 79.0& 80.5& 83.5& 79.7& 80.0& 84.7& 77.8& 64.2& 72.0& 48.7& 75.7& 46.8& 39.6& 74.9& 80.0& 60.2& 51.4& 40.7& 54.8& 45.1& 48.7& 67.1\\
GPT-4o            & 64    & 77.1& 80.5& 83.9& 76.5& 70.2& 83.8& 66.7& 62.2& 69.1& 49.2& 73.3& 41.2& 37.2& 43.6& 56.0& 44.5& 41.2& 38.4& 32.8& 56.9& 38.7& 60.2\\
Claude3.5 Sonnet  & 20    & 80.5& 77.3& 82.0& 81.7& 72.3& 75.4& 61.1& 61.8& 69.3& 43.1& 72.4& 31.6& 34.0& 32.8& 48.8& 36.8& 38.4& 34.8& 34.4& 64.7& 37.7& 57.7\\
\midrule
\multicolumn{24}{c}{Open-Source Video MLLMs} \\
\midrule
Qwen2-VL-7B           & 0.2-1fps & 75.2& 82.8& 73.2& 77.5& 68.3& 71.0& 72.2& 61.2& 61.5& 46.1& 69.0& 41.2& 22.0& 32.8& 43.6& 34.9& 31.2& 26.0& 39.6& 22.7& 31.7& 54.1
\\
MiniCPM-V-2.6-8B      & 32       & 71.9& 71.1& 77.9& 75.8& 64.6& 65.7& 70.4& 56.1& 62.3& 53.4& 67.4& 40.8& 24.0& 34.0& 41.2& 35.0& 34.0& 31.6& 41.9& 22.2& 35.0& 53.9
\\
InternVL-V2-8B        & 16       & 68.1& 60.9& 69.4& 77.1& 67.7& 62.9& 59.3& 53.3& 55.0& 56.5& 63.7& 37.6& 26.4& 37.2& 42.0& 35.8& 32.0& 31.2& 32.3& 40.9& 32.4& 51.4
\\
Kangaroo-7B           & 64       & 71.1& 84.4& 70.7& 73.2& 67.1& 61.7& 56.5& 55.7& 62.0& 38.9& 64.6& 37.6& 31.2& 28.8& 39.2& 34.2& 32.8& 26.4& 33.8& 16.0& 30.1& 51.1
\\
LongVA-7B             & 128      & 70.0& 63.3& 61.2& 70.9& 62.7& 59.5& 61.1& 53.7& 54.7& 34.7& 60.0& 39.6& 32.4& 28.0& 41.6& 35.4& 32.8& 29.6& 30.3& 15.9& 30.0& 48.7
\\
VILA-1.5-8B           & 14       & 53.7& 49.2& 71.0& 56.9& 53.4& 53.9& 54.6& 48.8& 50.1& 17.6& 52.3& 41.6& 26.4& 28.4& 36.0& 33.1& 26.8& 34.0& 23.2& 17.7& 27.4& 43.2
\\
Video-CCAM-14B         & 96       & 56.4& 57.8& 65.3& 62.8& 64.6& 51.4& 42.6& 48.0& 49.6& 31.6& 54.0& 33.6& 22.0& 28.4& 34.8& 29.7& 27.6& 24.4& 16.7& 22.7& 22.9& 42.5
\\
Video-LLaMA2-7B       & 32       & 55.9& 55.5& 57.4& 58.2& 52.8& 43.6& 39.8& 42.7& 45.6& 35.2& 49.5& 30.4& 32.4& 30.4& 36.0& 32.4& 24.8& 26.8& 18.7& 0.0& 21.9& 40.4
\\
\textbf{LLaVA-OV-7B }          & 32       & 80.4& 74.2& 76.0& 80.7& 72.7& 71.7& 67.6& 65.5& 65.7& 45.1& 71.1& 40.8& 37.2& 33.6& 44.8& 38.4& 35.6& 36.0& 27.3& 29.6& 32.7& 56.4\\
\midrule
\multicolumn{24}{c}{Streaming MLLMs} \\
\midrule
Flash-VStream-7B      & -    & 25.9& 43.6& 24.9& 23.9& 27.3& 13.1& 18.5& 25.2& 23.9& 48.7& 23.2& 25.9& 24.9& 25.6& 28.4& 26.0& 24.8& 25.2& 26.8& 2.0& 24.1& 24.0
\\
VideoLLM-online-8B    & 2fps & 39.1& 40.1& 34.5& 31.1& 46.0& 32.4& 31.5& 34.2& 42.5& 27.9& 36.0& 31.2& 26.5& 24.1& 32.0& 28.5& 24.2& 29.2& 30.8& 3.9& 26.6& 32.5
\\
Dispider-7B           & 1fps & 74.9& 75.5& 74.1& 73.1& 74.4& 59.9& 76.1& 62.9& 62.2& 45.8& 67.6& 35.5& 25.3& 38.6& 43.3& 35.7& 39.6& 27.7& 34.8& 25.3& 33.6& 53.1
\\
\textbf{ReKV-7B }              & 0.5fps & 74.4& 78.9& 78.6& 77.1& 68.3& 67.9& 67.6& 62.6& 64.3& 44.6& 69.1& 38.8& 24.8& 39.6& 46.4& 37.4& 31.2& 30.4& 30.4& 30.8& 30.7& 53.5\\
\midrule
\multicolumn{24}{c}{Ours} \\
\midrule
\textbf{StreamKV-7B ($\downarrow$60\%)} & 0.5fps & 74.7& 78.1& 87.7& 79.4& 70.8& 67.6& 70.4& 64.6& 64.0& 45.1& 71.0& 51.2& 39.7& 46.0& 68.4& 51.4& 45.6& 31.7& 36.0& 32.0& 36.4& 58.9
\\
\textbf{StreamKV-7B ($\downarrow$80\%)} & 0.5fps & 73.6& 77.3& 85.8& 77.8& 72.7& 64.8& 68.5& 63.4& 63.7& 44.6& 69.8& 48.4& 36.4& 45.6& 66.4& 49.3& 42.5& 31.2& 33.7& 31.9& 34.8& 57.4
\\
\textbf{StreamKV-7B ($\downarrow$90\%)} & 0.5fps & 73.8& 77.3& 85.9& 77.5& 73.3& 63.9& 69.4& 61.4& 63.2& 35.8& 68.8& 48.4& 36.4& 44.0& 66.1& 48.7& 43.6& 30.0& 33.2& 31.4& 34.6& 56.7\\
\bottomrule
\end{tabular}
\caption{Performance comparison on StreamingBench for Real-Time Visual Understanding, Omni-Source Understanding, and Contextual Understanding tasks.}
\label{StreamingBench}
\end{table*}

\subsection{Implementation details}
\noindent \textbf{Experimental setup}
We select the LLaVA-OneVision-Qwen2-7B-OV~\cite{li2024llavaonevision} as the baseline model due to its simplicity and strong performance.
All experiments are conducted on NVIDIA H20 GPUs (96G) with FP16 precision.
StreamKV processes video streams at 0.5 FPS, and the local window size is set to 15K, following the same experimental setup as ReKV. For dynamic partitioning, the minimum and maximum segment lengths are set to 4 and 64 frames, respectively, with a partitioning threshold of 0.99 employed to determine partitioning points. For KV-Cache retrieval, we set the number of retrieved frames to 8.

\subsection{Streaming Video Question Answering}

\begin{table*}[t]
\centering
\begin{tabular}{lcccccccccc}
\hline
Method & $\downarrow$0\% & $\downarrow$10\% & $\downarrow$20\% & $\downarrow$30\% & $\downarrow$40\% & $\downarrow$50\% & $\downarrow$60\% & $\downarrow$70\% & $\downarrow$80\% & $\downarrow$90\% \\
\hline
Uniform   & 60.49 & 59.57 & 58.94 & 58.57 & 57.65 & 57.32 & 56.33 & 55.56 & 53.02 & 51.41 \\
Semantic  & 61.20 & 60.18 & 60.18 & 59.36 & 59.54 & 59.07 & 58.89 & 58.11 & 57.43 & 56.72 \\
\hline
\end{tabular}
\caption{Ablation results on semantic partitioning across varying compression rates.}
\label{tab:semantic}
\end{table*}
\begin{table*}[t]
\centering
\begin{tabular}{lcccccccccc}
\hline
Method         & $\downarrow$0\% & $\downarrow$10\% & $\downarrow$20\% & $\downarrow$30\% & $\downarrow$40\% & $\downarrow$50\% & $\downarrow$60\% & $\downarrow$70\% & $\downarrow$80\% & $\downarrow$90\% \\
\hline
w/o. summary    & 60.52 & 59.46 & 59.25 & 58.44 & 57.95 & 57.42 & 56.21 & 54.66 & 54.65 & 53.85 \\
w/. summary     & 61.20 & 60.18 & 60.18 & 59.36 & 59.54 & 59.07 & 58.89 & 58.11 & 57.43 & 56.72 \\
\hline
\end{tabular}
\caption{Ablation results on the impact of the summary vector across varying compression rates.}
\label{tab:summary}
\end{table*}

We evaluate StreamKV on StreamingBench to assess its capability in streaming video understanding. Table~\ref{StreamingBench} presents comprehensive question-answering accuracies on StreamingBench, covering 18 subtasks organized into three categories: Real-Time Visual Understanding, Omni-Source Understanding, and Contextual Understanding. StreamKV significantly outperforms existing Online Video-LLMs, achieving state-of-the-art performance even when retaining only 10\% of the key-value pairs. Our experiments reveal several interesting observations:

1) In the Clips Summarization (CS) subtask, which involves summarizing the content of specific video segments, StreamKV with a 60\% compression ratio achieves a notably high accuracy of 87.7\%, representing an improvement of 9.1\% over ReKV and 11.7\% over the foundation LLaVA-OneVision model, and even outperforming all three proprietary MLLMs. This significant gain mainly stems from the semantic partitioning and summary vector mechanism, which effectively preserves key segment-level information, as well as our compression method, which effectively captures essential semantic elements within each segment.

2) For all four Omni-Source Understanding subtasks, StreamKV surpasses the performance of two proprietary MLLMs. These tasks evaluate a model’s ability to simultaneously process visual and audio content in video streams. While most existing Video LLMs cannot directly process audio, they instead infer visual scenes based on textual descriptions. 
StreamKV's superior performance in these subtasks demonstrates its ability to accurately capture fine-grained visual information over lengthy video stream. This advantage primarily stems from our layer-adaptive KV retrieval strategy, which enables the precise retrieval of the most relevant video KVs from the KV bank.

3) Similarly, StreamKV achieves superior performance on the Anomaly Context Understanding (ACU) task, which assesses the ability of MLLMs to detect and accurately identify unusual or unexpected events within a video stream. These results indicate that StreamKV can effectively capture subtle semantic changes and reliably recognize anomalies, thereby enabling precise understanding in dynamic and unpredictable environments.

\begin{table}[t]
\centering
\begin{tabular}{llccccc}
\hline
Com. & Ret. & $\downarrow$50\% & $\downarrow$60\% & $\downarrow$70\% & $\downarrow$80\% & $\downarrow$90\% \\
\hline
Uni. & Uni.   & 58.12 & 57.83 & 57.16 & 56.74 & 55.91 \\
Ada. & Uni.  & 58.49 & 58.44 & 57.53 & 57.11 & 56.35 \\
Uni. & Ada.  & 58.52 & 58.36 & 57.41 & 57.06 & 56.42 \\
Ada. & Ada & 59.07 & 58.89 & 58.11 & 57.43 & 56.72 \\
\hline
\end{tabular}
\caption{Ablation results for the Layer-Adaptive KV Selection Module across varying compression rates.}
\label{tab:adaptive}
\end{table}

\subsection{Ablations}
Unless otherwise specified, all of our ablation studies are conducted on StreamingBench, and the reported scores represent the overall performance across three task categories.

\noindent \textbf{Effectiveness of segment-level compression}
To verify that our semantic partitioning approach benefits the compression process, we compare the performance of compression on semantic segments versus uniform segments across various compression ratios. As shown in Table~\ref{tab:semantic}, compression performed on semantic segments consistently achieves superior performance across all compression ratios, indicating its greater effectiveness in preserving comprehensive visual information and contextual continuity.

\noindent \textbf{Effectiveness of the segment summary vector}
To assess the effectiveness of the summary vector, we evaluate StreamKV with and without incorporating the summary vector across various compression ratios. As illustrated in Table~\ref{tab:summary}, the performance of StreamKV is significantly better when the summary vector is included. This result suggests that the summary vector effectively preserves segment-level information and maintains the structural integrity of the video content, thereby enabling the model to generate more accurate answers.

\noindent \textbf{Effectiveness of Layer-Adaptive KV Selection Module}
We evaluate our proposed layer-adaptive selection module for both KV compression and retrieval. As shown in Table~\ref{tab:adaptive}, both compression and retrieval can use a uniform or adaptive selection budget across layers. The result demonstrates that the fully adaptive strategy consistently achieves the best performance across all compression ratios. Additionally, applying the adaptive strategy to either compression or retrieval alone also outperforms the fully uniform approach. 
This demonstrates the effectiveness of our layer-adaptive selection module.

\begin{figure}[t]
  \centering
  \includegraphics[width=0.36\textwidth]{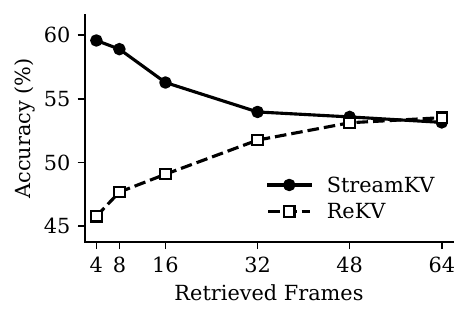}
  \caption{Accuracy versus Retrieved Frames Comparison for StreamKV and ReKV.}
  \label{fig:retrieved}
\end{figure}

\noindent \textbf{Number of retrieved frames}
We set the compression ratio at 60\%, i.e., discarding 60\% of the KV caches, and evaluate the impact of varying the number of retrieved frames. 
As illustrated in Figure~\ref{fig:retrieved}, we observe that increasing the number of retrieved frames in StreamKV actually leads to a decreased performance. 
This suggests that StreamKV is able to accurately retrieve the most relevant frames, and retrieving additional frames introduces extra irrelevant information~\cite{liu2023lostmiddlelanguagemodels}, which can in turn hinder the subsequent question answering process. 
In contrast, our experiments on ReKV show the opposite trend. ReKV needs to retrieve more frames to ensure that relevant information is included, due to its lower retrieval precision.
Moreover, retrieving fewer frames reduces computational overhead during the question answering stage, thereby accelerating inference. 
This demonstrates the effectiveness of our precise retrieval strategy and efficient inference.

\section{Conclusion}
In this paper, we introduced \textbf{StreamKV}, a training-free framework designed to address key challenges in streaming video understanding, including context preservation, long-term memory bottlenecks, and precise retrieval. By leveraging semantic partitioning and summary vectors, StreamKV effectively preserves the semantic continuity of video content. For efficient memory management, it employs a guidance prompt to identify and retain only the most informative KV caches within each segment. Furthermore, StreamKV proposes a Unified Layer-Adaptive KV Selection Module to further improve compression and retrieval. 
Extensive experiments on the StreamingBench benchmark demonstrate that StreamKV significantly outperforms existing online Video-LLMs in accuracy while substantially improving memory efficiency and reducing latency. Our work presents a practical and effective solution for building powerful and efficient Online Video-LLMs, paving the way for more robust real-world applications.

\clearpage
\bibliography{aaai2026}
\end{document}